\documentclass{article}
\usepackage{spconf,amsmath,graphicx,amssymb,url}
\usepackage{multirow}
\usepackage{makecell}

\newcommand{\etal}{\textit{et al.}}

\usepackage{amsmath}

\title{Spectral Graph Wavelet Transform as feature extractor for machine learning in neuroimaging}

\name{Yusuf Yigit Pilavci, Nicolas Farrugia}
\address{Lab-STICC - IMT Atlantique\\ Electronics Department\\ Brest, France}
\twoauthors
  {Yusuf Yigit Pilavci\textsuperscript{ a,b}}
	{\textsuperscript{ a}Politecnico di Milano\\
	Computer Science and Engineering \\ 
	Milano, Italy}
  {Nicolas Farrugia\textsuperscript{ b}}
    {\textsuperscript{ b}Lab-STICC - IMT Atlantique\\ Electronics Department\\ Brest, France}
\begin{document}
%
\maketitle
\begin{abstract}
Graph Signal Processing has become a very useful framework for signal operations and representations defined on irregular domains. Exploiting transformations that are defined on graph models can be highly beneficial when the graph encodes relationships between signals. In this work, we present the benefits of using Spectral Graph Wavelet Transform (SGWT) as a feature extractor for machine learning on brain graphs. First, we consider a synthetic regression problem in which the smooth graph signals are generated as input with additive noise, and the target is derived from the input without noise. This enables us to optimize the spectrum coverage using different wavelet shapes. Finally, we present the benefits obtained by SGWT on a functional Magnetic Resonance Imaging (fMRI) open dataset on human subjects, with several graphs and wavelet shapes, by demonstrating significant performance improvements compared to the state of the art. 
\end{abstract}
\begin{keywords}
graph signal processing, wavelets, neuroimaging, regression
\end{keywords}
\section{Introduction}
\label{sec:intro}

The emergence of Graph Signal Processing (GSP) is mostly due to the elegant and powerful analogy between graph laplacian eigenvectors and classical Fourier analysis~\cite{shuman_emerging_2013}. As such, application domains involving graphs to model mutltivariate dependencies are naturally adapted to the GSP framework. In particular, recent studies in neuroimaging have leveraged the use of graph theory to study brain networks, giving rise to the field of network neuroscience~\cite{bassett2017network}. 
However, the majority of network neuroscience studies have focused on analyzing the properties of the graph itself (e.g. graph theoretical descriptors of white matter connectivity~\cite{bassett2017network}), rather than signals of brain activity on a brain graph. In turn, most studies analyzing brain signals relies on the use of massively univariate statistics, analyzing each brain region independently~\cite{friston1994statistical}. Recent methodological efforts push towards the use of machine learning as multivariate methods able to capture whole brain dependencies of neural activity~\cite{Varoquaux2014}. 

Our aim with the present paper is to demonstrate the potential of simultaneously using brain connectivity and signals from neuroimaging data to boost performance of machine learning tasks, by leveraging the GSP framework. In particular, we aim at testing the benefits of using SGWT as a feature extractor for machine learning on neuroimaging data. 

The remainder of the paper is organized as follows. In section~\ref{sec:relatedwork} we will present previous studies related to SGWT and GSP applied to machine learning in neuroimaging. In section~\ref{sec:methods} we will present our experimental setup, the formulation of SGWT that will be used in this paper and the procedures used to build our graph models. In section~\ref{sec:results} we present experiments and results on a synthetic regression problem and fMRI signals. Finally, we conclude in section~\ref{sec:conclusion}.

\section{Related work}
\label{sec:relatedwork}
With the development of GSP, frameworks based on wavelets on graphs have become promising, due to their over-complete analysis in both localization on graph nodes and different scales with respect to the spectrum of the graph. To obtain such analysis on irregular domains such as graphs, several formulations of graph wavelets have been proposed~\cite{narang2012perfect,crovella2003graph,HAMMOND2011129}. For example, Narang reports a compact comparison of several formulations in terms of significant properties such as perfect reconstruction or orthogonality~\cite{narang2012perfect}. In this paper, we use Spectral Graph Wavelet Transform (SGWT) introduced in~\cite{HAMMOND2011129}. Basically, SGWT exploits the Fourier transform analogy that is defined on graphs~\cite{shuman_emerging_2013} and defines wavelet kernel functions in the spectral domain. 

While there is a growing literature applying GSP to neuroscientific questions, (for a review, see~\cite{huang2017graph}), most studies use GSP to derive descriptors (such as alignement of functional signals to the underlying graph~\cite{medaglia2018functional}) that are further analyzed using inference based statistics (e.g. \cite{Smith_2017}). For example, Leonardi~\cite{leonardi2011wavelet} shows the statistical relevance of SGWT coefficients in different scales with varying experimental conditions in an fMRI paradigm.

However, studies applying GSP for machine learning in neuroimaging are relatively scarce. In~\cite{rui2017simultaneous}, the authors defined a low rank, dimensionality reduction approximation and recovers the underlying graph, showing performance improvements when using the learnt approximation as features for supervised classification. In \cite{menoret2017evaluating}, the authors used GSP as dimensionality reduction and feature extraction in a supervised learning setting with fMRI. The authors show performance improvements in both simulated cases and real data. A recent contributions uses GSP to extract features for brain computer interfaces based of near-infrared spectroscopy brain signals~\cite{petrantonakis2018single}, showing significant performance improvements compared to previous work on the same dataset. Nevertheless, to the best of our knowledge, there is no published study exploiting SGWT as feature extractor for machine learning in neuroimaging. 

\section{Methods}
\label{sec:methods}
\subsection{Graph Wavelet Transform}
    Using graph wavelet transform enables multi-scale signal representations adapted to the underlying graphs, which in turn facilitate the detection of abnormal changes or discontinuities in the original domain, and ease the interpretation of signals in both localization and frequency. With these motivations, we adopt the formulation of SGWT~\cite{HAMMOND2011129}.
    
    As in classical signal processing, wavelet functions are defined as $\psi_{s,a}(x)$ for different scales $s$, and translations $a$. Graph Wavelets can be also interpreted in the Fourier domain as: 
    \begin{equation}
        \psi_{s,a}(x) = \frac{1}{2\pi}\int_{-\infty}^{\infty}\hat{\psi}(s\omega)e^{-j\omega a}e^{j\omega x}d\omega
        \label{eq:waveletfourier}
    \end{equation}
    where $\hat{\psi}(s\omega)$ is the Fourier transform of the scaled wavelet, $e^{-j\omega a}$ is the Fourier transform of the spatial translation, (which can be seen as the Fourier Transform of a delta localized at $a$) and $e^{j\omega x} $ is the Fourier basis function. These terms will be addressed to corresponding terms in graph wavelet transform definition. 
    
    Let an undirected, weighted graph $G$, with a vertex set $V$ such as $|V| = N$, and weight matrix $\mathcal{\textbf{W}\in\mathbb{R^N}}$. $\mathcal{\textbf{W}}$ is a symmetric matrix of weights ${w}_{ij}$. The Laplacian operator of $G$ is defined as follows: 
    \begin{equation}
        L = D - W
    \end{equation}
    where $D$ is a diagonal matrix such as $D_{ii} = \sum_{j}|{w}_{ij}|$. Here, weights ${w}_{ij}$ are allowed to be negative, and we deal with issues related to semi-positiveness of the $L$ by using the absolute diagonal degree matrix~\cite{leonardi2011wavelet}. Also, the normalized Laplacian is defined as $L_{norm} = D^{-1/2} L D^{-1/2}$. Analogous to classical signal processing, eigenvectors $\textbf{U}$ and eigenvalues of the Laplacian matrix $\textbf{L}$ correspond to the Fourier basis and frequency values, respectively. A graph signal $f$ is a vector in~$\mathbb{R^N}$ defined in the vertex domain~\cite{shuman_emerging_2013}, and spectral filtering operations $f$ are done through Graph Fourier Transform, defined by $\hat{f} = \textbf{U}^Tf$. 
    
    By interpreting scaling and translation operations, as in equation~\ref{eq:waveletfourier}, it is possible to define graph wavelets as follows: 
    \begin{equation}
        \psi_{s,a} = \sum_{n=1}^N{g(s\lambda_n)\hat{\delta}(n)u_n}
    \end{equation}
    where $g$ is a band-pass kernel defined in the spectral domain,  corresponding to the Fourier transform of the wavelet at scale $s$ in equation~\ref{eq:waveletfourier}, and $\hat{\delta}(n)$ is the Graph Fourier transform of $\delta$ localized at $a$, corresponding to the spatial translation component in equation~\ref{eq:waveletfourier}. Finally, $u_n$ are the columns of $\textbf{U}$, eigenvectors of $\textbf{L}$. The complete frame for the wavelet transform is computed by adding a low-pass filter, $h(\lambda)$, also called the scaling function. As a result, the obtained transform covers all parts of the graph spectrum. The final transformation for a graph signal $f$ is the following inner product: 
    \begin{equation}
        W_f(s,a) = <\psi_{s,a},f>
    \end{equation}
    

    \subsection{Graph Wavelet Kernel Design}
    The informative property of SGWT as signal representation is highly related to the selection of the scaling and wavelet functions, $h(\lambda)$ and $g(\lambda)$. Such a choice impacts the stability of reconstruction of the original graph signal. As reported in~\cite{HAMMOND2011129} and~\cite{Shuman2015SpectrumAdaptedTG}, a wavelet frame would be tight if the sum of squares of all kernels remains constant through the spectrum (Parseval identity), a necessary condition for perfect reconstruction of a signal. However, it is possible to relax this constraint by accepting some variation, resulting in more freedom of kernel selection (e.g. using cubic splines). 
    Another aspect of kernel design is spectrum coverage. For a graph model, the spectrum is defined by the eigenvalues of $\textbf{L}$, and any continuous function that is defined in the spectral domain (as a function of $\lambda$) is only evaluated at those eigenvalues, $\lambda_n ,$~\cite{Shuman2015SpectrumAdaptedTG}. Therefore, one should examine the eigenvalue locations through the spectrum when designing kernel functions. 
    
\begin{figure}
  \centering
    \includegraphics[width=0.45\textwidth]{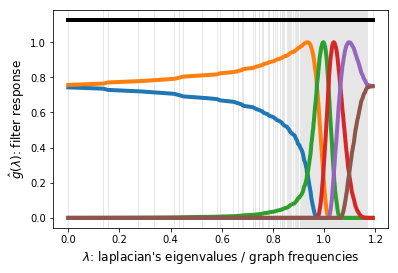}
  \caption{Warped Translate Wavelet Kernels on Spectral Domain, in the case of the KNN-Correlation Brain Graph. Vertical lines depict placement of eigenvalues.}
    \label{fig:warpedtranslate}
\end{figure}

    Shuman \etal provides tight and spectrum adapted wavelet kernels, called Warped Translates. In~\cite{Shuman2015SpectrumAdaptedTG}, a tight frame is generated with a special function:
    \begin{equation}
        q(\lambda) := \sum_{k=0}^{K}a_k\cos \left({2\pi k(\omega(\lambda)-\frac{1}{2})}\right)\mathbf{1}_{0\leq \lambda<1}
    \end{equation}
    and its translated versions in the graph spectrum, 
    $q\left(\omega(\lambda)-\frac{m}{R}\right) $ where $\sum_{k=1}^K(-1)^ka_k=0$, and $\omega$ is a non-decreasing warping function that modifies the kernels' behaviour on the spectrum. For example, choosing a warping function as the approximated cumulative distribution function of eigenvalues concentrates the kernels in ranges in which the eigenvalues are densely placed (see Fig.~\ref{fig:warpedtranslate}). In~\cite{Shuman2015SpectrumAdaptedTG}, the superior discriminatory power of warped translates is clearly demonstrated, because different parts of the evaluated spectrum are perfectly covered and segmented by different kernels.  
    
    \subsection{Synthetic Graph Signals and Regression Problem}
    \label{sec:synth}
    In this work, synthetic signals are generated for getting more understanding on the behavior of SGWT in a regression problem, in the case where the signals are smooth on the graph. The workflow that generates the input and output starts with creating a Erdos Renyi graph (with a probability of edge presence of $p=0.1$), $G$ with $N$ nodes and associated weight matrix $\mathbf{W}$. A diffusion operator is computed using a lazy random walk, $\mathbf{A} = ( {I} + D^{-1}{W}) $. Secondly, we generate a random signal matrix $\mathbf{R}$ of size $M \times N$, where $M$ is number of samples and $N$ is number of features,  and the generated set of signals are diffused by multiplying it by the diffusion operator, $\mathbf{A}$, such as the diffused signals are $\mathbf{\hat{R}=RA}$. The third step is to generate a random vector of regression weights $\beta$, and compute the nonlinear function output as $y = \log(\beta^T \hat{R})$. Finally, zero-mean Gaussian noise is added on the graph signals, so the final observed signals are $\mathbf{X = \hat{R} + \mathcal{N}(0,{\sigma}^2)}$, with $\sigma=0.1$. With this procedure, we are able to obtain graph signals,  and also generate a problem that cannot be easily solved by linear regressors. Those signals are smooth on the graph likewise the FMRI signals on the brain nodes. 

    
    \subsection{fMRI Datasets}
    We consider here an open dataset of fMRI data on human subjects who rated pictures with emotional content~\cite{chang2015sensitive}. We fetched statistical maps of whole brain activity during single trials for each subject, from neurovault~\cite{gorgolewski2015neurovault} (collection number 1964). The supervised learning task consists in predicting the rating given by the subject from brain maps.
    As this dataset didn't include connectivity data that could be used to compute subject-specific graphs, we estimated average brain graphs from resting state fMRI data (i.e. spontaneous fluctuations of the brain at rest) from 158 subjects of an open resting-state dataset~\cite{mendes2017functional}. We used the preprocessed  resting-state data, described here \url{https://neuroanatomyandconnectivity.github.io/opendata/}. As both datasets were spatially normalized in the standard MNI space, we defined regions of interest (ROI) to enable the correspondance between the graph, defined on the resting-state dataset, and the signals from the emotional rating dataset. Therefore, we parcellated all resting-state time-series and brain maps on 523 non-overlapping ROIs from the finest scale of BASC atlas (444 networks)~\cite{bellec_multi-level_2010}.  
    
\begin{table*}

\centering

 \begin{tabular}{||c || c | c | c || c | c | c || c | c ||} 
 \hline
 \multirow{2}{*}{\makecell{Data \\ Repr.}} & \multicolumn{3}{c||}{CV Results} & \multicolumn{3}{c||}{Test Results}& \multicolumn{2}{c||}{Transform Properties}  \\ \cline {2-9}
 & RMSE & $R^2$-score & Pearson  & RMSE & $R^2$-score & Pearson & Graph & Kernel  \\ [0.5ex] 
 \hline\hline
 Original  & 1.080 $\pm$ 0.028 & 0.204 $\pm$ 0.073 & 0.810 $\pm$ 0.018   & 1.036 & 0.451 & 0.693 & NA & NA   \\
 ROI  &	1.054 $\pm$ 0.031 & 0.231 $\pm$	0.079 & 0.827$\pm$ 0.017 & 1.022 &	0.466 &  0.701	& NA & NA\\
 SGWT-1  & 1.047 $\pm$ 0.029 	& 0.234$\pm$ 0.082 & 0.832 $\pm$ 	0.015	 &  \textbf{ 0.983} & \textbf{0.506 }&  0.725 & Corr. &   Warped   \\
 
 SGWT-2  & 	1.067$\pm$	0.030 &	0.205 $\pm$	0.085  &	0.840 $\pm$ 0.014 &   0.987 & 0.502 & \textbf{0.737} & KNN Corr. &  Warped   \\
 
 SGWT-3  & 1.056$\pm$	0.034 	& 	0.205$\pm$ 0.085 &  0.839$\pm$0.014	 &   0.988 & 0.500 & 0.722 & KNN Cov & Meyer \\
 
 SGWT-4  &  \textbf{1.033} $\pm$	\textbf{0.030} 	& 	\textbf{0.245} $\pm$ \textbf{0.082}  & 0.828$\pm$	0.016 &  0.990 & 0.498 & 0.720 & Kalofolias & Cubic Spline \\
 
 SGWT-5  & 1.062 $\pm$	0.031 	& 0.196 $\pm$	0.107 & \textbf{0.843} $\pm$	\textbf{0.014}	 &   0.991 & 0.487 & 0.730 & KNN Corr. & Iter. Sinus \\

\hline
 \end{tabular}
\caption{Results for the fMRI Dataset}
\label{tab:pinesresult}
 
\end{table*}

\begin{table}[h!]

\centering

 \begin{tabular}{||c || c | c ||} 
 \hline
 
Wavelet Kernels & MSE & $R^2$-score \\
 \hline
Cubic Spline  & 1514.45e-05 
&   0.392438 
\\
Meyer & 1518.24e-05  
&  0.390937  
\\ 
Iterated sinusoidal & 1515.33e-05  
& 0.392079 
\\ 
\textbf{Warped Translate} & \textbf{1507.40e-05} 
& \textbf{0.395246} 
\\
 \hline
No Wavelet & 1533.48e-05  
& 0.384859 
\\
 \hline 
 \end{tabular}
\caption{Results for the Synthetic dataset}
 \label{tab:synth}

\end{table}
    \subsection{Graph Construction}
We explain here the different graph inference methods applied on the parcellated resting state time-series. The weight matrix, \textbf{W} formulations are as follows:  
\begin{itemize}
\item Functional Connectivity with Covariance or Correlation Graph: We derive graphs from temporal covariance and correlation. As variant of these graphs, thresholded and binary versions of these adjacency matrices are also defined.
    \item KNN Covariance/Correlation Graph: Another variant to previous graphs is generated by limiting the numbers of neighbors per node. For each node, the K strongest edges are kept and rest are set to a weight of 0. For the symmetry, when $\textbf{W}_{ij}$ is set with nonzero edge, same value is kept in $\textbf{W}_{ji}$. Binary versions are also generated by setting nonzero values to 1. 
      
    \item Semi-Local Graph: Another strategy that exploits both the geometrical structure and functional connectivity is the semi-local graph, as tested in ~\cite{menoret2017evaluating}. The semi-local graph uses a threshold (set to keep the graph connected) on the euclidean distance between baricenters of nodes, and sets connection weights corresponding to long distances to 0. 
    
    \item Kalofolias Graph: We used the method from Kalofolias~\cite{kalofolias2016learn}, which relies on defining a smoothness prior to infer the graph from the data. 
    
\end{itemize}

    \subsection{Dimensionality Reduction and Regression}
    For the synthetic dataset, generated inputs $X$ are first transformed using SGWT, followed by dimensionality reduction using 100-best feature selection. Performances of linear regression predicting $y$ are compared with and without SGWT for different kernels. For the fMRI dataset, we reproduced the methods from the original paper~\cite{chang2015sensitive} to be able to compare results. A PCA with 121 components is used for dimensionality reduction, and regression is performed with Lasso. Careful cross-validation (CV) is performed over subjects and regularization parameter for LASSO, using leave-one-subject-out, enabling reliable generalization. We compared the original pipeline that uses no parcellation (original signal), parcellated signals on BASC ROI, and SGWT representations.

\section{Results}
\label{sec:results}
For the evaluation of results on the synthetic dataset, $R^2$-score and Mean Square Error (MSE) are used, whereas for the fMRI dataset, Root mean square error (RMSE) and Pearson correlation are employed , to be consistent with~\cite{chang2015sensitive}. Variability in generalization is reported using the average and the standard error of the mean (SE) over CV folds for each score. In addition, we report results on a generalization test set using a predefined, separate hold-out test dataset as explained in ~\cite{chang2015sensitive}.

Table \ref{tab:pinesresult} presents results obtained for the fMRI dataset. Result for SGWT are picked from the set of best results for different graphs and wavelet function. In Table \ref{tab:pinesresult}, the best five results for SGWT are reported and these scores stands for the results of different graph and kernel selections. 
SGWT provides significant performance gains when compared to using only parcellated ROI signals, or original signals with no parcellation. In particular, warped translate have a better potential for generalization to the test dataset. We suggest that SGWT enables an efficient exploitation of underlying multivariate dependencies, using spectrum-adapted wavelet kernels on a brain 
graph. While the magnitude of performance gain is strongly dependent to underlying information inferred by the graph, warped translates are able to extract more informative features in terms of regression problems. 

\begin{figure}
  \centering
    \includegraphics[width=0.5\textwidth]{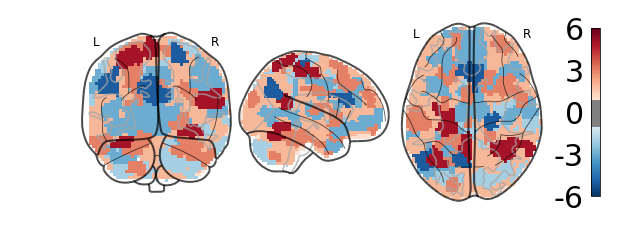}
  \caption{Significant Scale-Localization map on brain. Positive and negative weights are denoted with $\pm$ sign.}
    \label{fig:allscalenegposbeta}
\end{figure}

Comparisons for the synthetic dataset are reported in Table \ref{tab:synth}. The experiment for this analysis is repeated 500 times with randomly generated graphs with 500 nodes, and average scores are interpreted. As indicated by Table \ref{tab:synth}, SGWT, and in particular warped translate kernels are able to generate features that are significantly more informative to solve the regression problem. This indicates that the regression problem considerably benefits from using discriminative, spectrum-adaptive coverage of kernels w.r.t. others. 

We depict in Fig. \ref{fig:allscalenegposbeta} the different scale and spatial localization of the largest LASSO coefficients that solve the regression problem. Conceptually, this representation shows the most important graph scales for each localization in brain. Negative and positive weights are separately examined and the largest contribution for each node is denoted with its sign and scale in Fig.\ref{fig:allscalenegposbeta}. 

\section{Conclusion}
\label{sec:conclusion}
In this contribution, we have demonstrated the potential of combining SGWT and machine learning using synthetic data and fMRI data from open datasets. A key point of the proposed approach is to rely on Warped Translated kernels for wavelet definitions, which optimizes spectral coverage of SGWT. We showed that using features from SGWT can boost performance in a challenging regression task on neuroimaging data. In future work, we plan to better examine how the estimated features can be used to enhance interpretability of the trained models. Another perspective is to define new models based on dynamic graphs in order to fully exploit the temporal dimension of brain activity, instead of relying solely on spatial maps as graph signals.

\vfill\pagebreak

\bibliographystyle{IEEEbib}
\bibliography{strings,refs}

\end{document}